\documentclass[letterpaper]{article}

\usepackage[T1]{fontenc}

\usepackage{geometry}
\usepackage{setspace}

\usepackage[style = chem-acs]{biblatex}
\usepackage{graphicx}
\usepackage{float}
\newfloat{scheme}{htbp}{los}
\floatname{scheme}{Scheme}
\floatname{chart}{Chart}
\newfloat{graph}{htbp}{loh}
\usepackage{algpseudocode}
\usepackage[ruled,vlined]{algorithm2e}
\usepackage{amsmath}
\usepackage{amssymb}
\usepackage{array}
\usepackage{xcolor}
\usepackage{hyperref}
\usepackage{chemformula} 
\usepackage[version = 4]{mhchem} 

\newcolumntype{C}[1]{>{\centering\arraybackslash}m{#1}}

\usepackage{authblk}
\author[1,+]{Sabrina Islam}
\author[1,+]{Md. Atiqur Rahman}
\author[1]{Md. Bakhtiar Hasan}
\author[1]{*Md. Hasanul Kabir}
\affil[1]{Computer Science and Engineering, Islamic University of Technology, Bangladesh}
\affil[+]{These authors contributed equally}

\title{Rep3Net: Unified Multimodal Molecular Representations for Inhibitor Prediction}
\date{*hasanul@iut-dhaka.edu}

\begin{document}

\maketitle

\begin{abstract}
Accurate prediction of compound potency accelerates early-stage drug discovery by prioritizing candidates for experimental testing. However, many Quantitative Structure-Activity Relationship (QSAR) approaches for this prediction are constrained by their choice of molecular representation: handcrafted descriptors capture global properties but miss local topology, graph neural networks encode structure but often lack broader chemical context, and SMILES-based language models provide contextual patterns learned from large corpora but are seldom combined with structural features. To exploit these complementary signals, we introduce Rep3Net, a unified multimodal architecture that fuses RDKit molecular descriptors, graph-derived features from a residual graph-convolutional backbone, and ChemBERTa SMILES embeddings. We evaluate Rep3Net on a curated ChEMBL subset for Human PARP1 using fivefold cross validation. Rep3Net attains an MSE of $0.83\pm0.06$, RMSE of $0.91\pm0.03$, $R^{2}=0.43\pm0.01$, and yields Pearson and Spearman correlations of $0.66\pm0.01$ and $0.67\pm0.01$, respectively, substantially improving over several strong GNN baselines. In addition, Rep3Net achieves a favorable latency-to-parameter trade-off thanks to a single-layer GCN backbone and parallel frozen encoders. Ablations show that graph topology, ChemBERTa semantics, and handcrafted descriptors each contribute complementary information, with full fusion providing the largest error reduction. These results demonstrate that multimodal representation fusion can improve potency prediction for PARP1 and provide a scalable framework for virtual screening in early-stage drug discovery.

\end{abstract}

\section*{Keywords}

Molecular Representation, Poly [ADP-ribose] polymerase 1 (PARP1), Inhibitor prediction, Drug Discovery, ChemBERTa, GNN.



\section{Introduction}
Efficient identification of small-molecule inhibitors against protein targets is a cornerstone of modern drug discovery: inhibitors can selectively modulate disease-relevant proteins, enabling targeted therapeutic strategies while reducing downstream experimental burden. Traditional experimental pipelines: high-throughput screening, hit-to-lead optimization, and preclinical evaluation, are resource intensive and time-consuming \cite{khambhati2024current, intuitionlabs2023drug, hughes2011principles, gillespie2004hit}. As a consequence, computational screening methods that predict bioactivities from molecular structure have become indispensable in early-stage discovery. In particular, Quantitative Structure-Activity Relationship (QSAR) modeling using machine learning and deep learning offers a practical route to prioritize compounds and focus experimental effort \cite{cherkasov2014qsar}.

Poly(ADP-ribose) polymerase 1 (PARP1) is a DNA damage-sensing enzyme whose catalytic PARylation promotes repair of single-strand breaks and supports tumour cell survival \cite{morales2014review,pascal2015rise,teloni2015readers}. Pharmacological inhibition of PARP1 induces synthetic lethality in tumours with defects in homologous recombination (for example, BRCA1/2 mutations) and has therefore emerged as an important therapeutic approach \cite{lord2017parp,kanev2024parp1}. Despite clinical successes, discovery and optimization of novel PARP1 inhibitors remain costly and slow; accordingly, computational prediction of inhibitory potency (commonly reported as $IC_{50}$, the concentration producing 50\% inhibition \cite{sebaugh2011guidelines}) is an attractive strategy to increase throughput and guide experimental validation \cite{mateo2019decade,lafargue2019exploring,vamathevan2019applications}.

A critical determinant of predictive performance in QSAR and related models is the molecular representation. Traditional fingerprint- and descriptor-based approaches capture global physicochemical properties \cite{david2020molecular,yang2022concepts}, Graph Neural Networks (GNNs) encode local and topological atom-bond relationships \cite{wang2025graph,xing2023graph}, and sequence-style models operating on SMILES strings can extract syntactic and contextual patterns \cite{liyaqat2024advancements,wigh2022reveiw}. Hybrid and multimodal approaches that combine representations have shown promise in several screening contexts. For example, message-passing and graph pipelines have been used for staged selection in PARP-family screens \cite{chen5571319ai}, and multimodal fusions of fingerprints, graph encoders, 1D-CNN SMILES encoders, or transformer-style sequence models have been explored for other targets \cite{10.1117/12.3044287,ijms26041681,ai2022multi}. However, most prior work trains all components from scratch and therefore requires substantial labeled data and compute, limiting applicability for targets where experimental labels are scarce.

Recent advances in pretrained molecular language models (e.g., SMILES or SELFIES masked-language models) produce contextual embeddings that encode chemical patterns learned from large unlabeled corpora. The opportunity to combine such pretrained SMILES embeddings with classical descriptors and graph representations, thereby uniting global, topological, and contextual information, remains under-explored for inhibitor potency prediction. We hypothesize that a fusion of these complementary modalities can improve both accuracy and sample efficiency relative to single-modality models.

To test this hypothesis, we introduce Rep3Net, a multi-branch architecture that (i) ingests three complementary representations for each molecule: expert molecular descriptors (global physicochemical features), a GNN-processed molecular graph (local/topological features), and contextual SMILES embeddings from a pretrained masked-language model (semantic/contextual features); (ii) learns modality-specific encodings and fuses them into a joint representation via end-to-end training; and (iii) employs a residual graph-convolutional backbone to deepen representation capacity while stabilizing optimization. We evaluate Rep3Net on PARP1 $pIC_{50}$ prediction and compare it with diverse graph-based and hybrid baselines.

This work makes three primary contributions:
\begin{itemize}
	\item We propose a multimodal representation strategy that unifies global molecular descriptors, learned graph features, and pretrained SMILES embeddings for inhibitor-potency prediction.
	\item We present Rep3Net, an end-to-end trainable architecture built around a residual graph-convolutional backbone and modality-specific encoders that support joint learning and efficient fusion of complementary features.
	\item We provide a comprehensive empirical evaluation and ablation study that quantifies how each modality contributes to predictive performance and sample efficiency for PARP1 $pIC_{50}$ prediction.
\end{itemize}

\section{Background Study}
Computational drug discovery is shaped strongly by the choice of molecular representation, and a long-standing axis of work contrasts engineered descriptors and fingerprints with learned structural or sequence-based embeddings. Classical descriptor- and fingerprint-based QSAR pipelines (typically built from standardized toolkits such as RDKit) remain widely used because they are interpretable, inexpensive to compute, and provide strong baselines in low-data regimes when paired with robust learners (e.g., Random Forests, XGBoost, SVM) \cite{landrum2013rdkit}. Nevertheless, descriptor-based approaches depend on manual feature design and can miss higher-order or context-dependent chemical signals; they are also vulnerable to biased, noisy, or sparse experimental labels because their fixed feature set limits adaptability to new chemotypes \cite{cherkasov2014qsar,wu2018moleculenet,mayr2018large}.

Representation learning on molecular structure and sequence has emerged to address these expressivity limits. Graph Neural Networks (GNNs) and message-passing architectures directly model atom-bond topology and relational patterns, producing permutation-invariant representations that capture local chemistry and aspects of global topology \cite{gilmer2017neuralmessagepassingquantum,zhou2020graph}. GNNs have advanced performance on many property-prediction tasks, but they commonly require larger labeled datasets, are sensitive to architectural and training choices, and can struggle to capture long-range or sequence-like contextual regularities that are naturally expressed in SMILES strings \cite{yang2019analyzing}. Parallel to graph methods, SMILES-based sequence models, especially transformer-style encoders pretrained with masked-language objectives, learn rich contextual embeddings from large unlabeled corpora and often transfer well to downstream tasks in data-scarce settings \cite{schwaller2019molecular,mswahili2024transformer,honda2019smiles}. The central open question is how sequence-derived contextual knowledge complements the structural fidelity of graphs and the global summaries provided by descriptors.

Hybrid and multimodal fusion strategies aim to combine these complementary strengths, and a number of studies report gains from two-modality fusions or staged pipelines (for example, fingerprints + GNNs or SMILES CNNs + GNNs) in diverse screening tasks \cite{10.1117/12.3044287,ijms26041681,chen5571319ai,ai2022multi}. Yet existing multimodal work shows three recurrent limitations: (a) many systems train all components from scratch, increasing data and compute requirements; (b) fusion strategies are heterogeneous and lack standard, reproducible best practices; and (c) empirical evaluations often omit rigorous modality-wise ablations that would quantify each representation's marginal contribution to performance and sample efficiency. These methodological gaps matter for therapeutically important targets such as PARP1, where labeled $IC_{50}$ data are costly and limited; most published computational PARP1 studies still emphasize descriptor-based QSAR pipelines \cite{shahab2025machine, gomatam2024improved, aldakheel2025integrating}, leaving open whether triple-modality fusion and reuse of pretrained SMILES embeddings can materially improve predictive accuracy and robustness.  

In this work, we propose Rep3Net to address these gaps: it combines expert descriptors, a residual graph-convolutional backbone, and pretrained SMILES embeddings in a unified, end-to-end framework and emphasizes systematic ablation to quantify modality contributions and sample efficiency for PARP1 $pIC_{50}$ prediction. By explicitly evaluating joint training versus reuse of pretrained components and reporting modality-wise ablations, our study aims to provide actionable guidance for future multimodal QSAR systems and for practitioners working on targets with limited experimental labels.

\section{Methods}

\subsection{Dataset}
ChEMBL database \cite{zdrazil2024chembl} was used as the source of bioactivity measurements for Human PARP1 (CHEMBL3105). The ChEMBL database is a comprehensive, high-quality resource that contains detailed information on bioactive compounds and their interactions with biological targets. It includes data on compound activity, chemical structure, pharmacological properties, and therapeutic applications, supporting research in drug discovery and molecular biology. ChEMBL is maintained by the European Bioinformatics Institute and provides access to curated data from experimental studies, enabling researchers to explore the relationships between chemical structure, bioactivity, and disease outcomes. This dataset serves as an invaluable tool for researchers engaged in computational drug discovery and molecular pharmacology.

From the database, we retrieved 4,802 compound records associated with $IC_{50}$ measurements. After applying the filtering and aggregation steps (detailed below), 3,356 unique samples remained and were used in model development and evaluation. For model evaluation, we performed 5-fold cross validation (with a random seed of 42). Within each fold, the data were partitioned into train, validation, and test sets in a 75:5:20 ratio, yielding around 2,500 training samples per fold.

\subsubsection{Preprocessing}

The dataset contains $IC_{50}$ measurements reported in nanomolar (nM) units and SMILES strings that represent molecular structure. To ensure consistency for the regression task and to limit noise from non-quantitative measurements, the following sequential filtering and aggregation steps were applied:

\begin{enumerate}
    \item Records reporting approximate or censored $IC_{50}$ values (e.g., ``$>$'', ``$<$'', or approximate annotations) were excluded, retaining only entries with explicit numerical $IC_{50}$ values.
    \item Entries lacking a valid SMILES string were removed as SMILES are required for the ChemBERTa and graph-based representations.
    \item Multiple records corresponding to the same compound were identified using canonical SMILES. Duplicates were aggregated by taking the median of the $IC_{50}$ values (aggregation performed on the $IC_{50}$ values as reported, in nM) to reduce the influence of experimental variation.
    \item All $IC_{50}$ values were retained in their reported units (nM) for bookkeeping and aggregation. However, regression on raw $IC_{50}$ values spans several orders of magnitude and can lead to large initial losses and unstable optimization. We therefore decided to convert $IC_{50}$ values to $pIC_{50}$ for regression. For this computation, $IC_{50}$ values were converted to molar units (M) by multiplying $10^{-9}$ prior to applying the following logarithmic transform:
    \begin{equation}
  pIC_{50} = -\log_{10}(IC_{50})
  \label{eqn:pic50}
\end{equation}
The resulting $pIC_{50}$ targets were standardized using \texttt{StandardScaler} \cite{pedregosa2011scikit} prior to training for a more stable training.
\end{enumerate}

\begin{figure}[htb]
        \centering
        \includegraphics[width=0.7\textwidth]{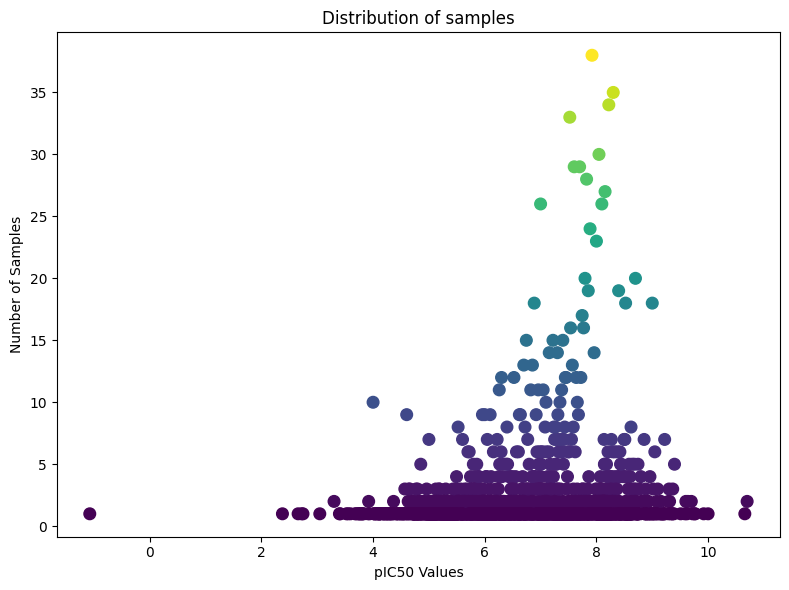}
\caption{Distribution of samples based on $pIC_{50}$ values. Each point represents the number of samples observed at a specific $pIC_{50}$ value, highlighting the concentration of compounds in the mid-range and the relative sparsity at extreme values.}
\label{figure: dataset}
\end{figure}

After these steps, 3,356 unique compounds remained.  The sample distribution of $pIC_{50}$ values is shown in \autoref{figure: dataset}.

\subsection{Input Representation}
In our architecture, each compound is represented by three complementary modalities: (i) hand-crafted molecular descriptors computed with RDKit for capturing global physicochemical properties, (ii) contextual SMILES embeddings from a masked language model (ChemBERTa) to extract contextual substructure and cheminformatic semantics, and (iii) a graph representation processed by a graph neural network to identify local topological and relational information. These representations are concatenated and forwarded to a downstream regressor.

\subsubsection{Molecular Descriptors}
Molecular descriptors provide quantitative representations of chemically relevant, hand-crafted properties of compounds \cite{grisoni2018impact}. By translating molecular structures into numerical values, these descriptors enable computational models to systematically characterize, compare, and analyze chemical compounds. As a result, molecular descriptors constitute fundamental feature inputs for machine learning and deep learning approaches in drug discovery, molecular property prediction, and chemical similarity analysis.

We computed an initial set of 217 molecular descriptors for each compound using RDKit. They were filtered to reduce redundancy and improve numerical condition:

\begin{itemize}
    \item Descriptors with variance lower than 0.01 across the dataset were removed, as they do not contribute much to decision boundaries.
    \item For pairs of descriptors with Pearson Correlation Coefficient greater than 0.9, one descriptor was removed as they essentially capture the same underlying pattern in the data, and retaining both does not contribute additional information. We removed one of the pairs using an upper-triangular selection procedure; the earlier-appearing feature was retained.
    \item The remaining descriptors were normalized to ensure that descriptors with large numerical ranges do not dominate those with smaller ranges. This improves the performance, stability, and convergence of our framework.
\end{itemize}

These steps reduced the descriptor set to 134 features per compound, lowering dimensionality and reducing the risk of overfitting while preserving chemically relevant information.


\subsubsection{Embeddings from MLM}

While molecular descriptors provide numerical representations of fundamental chemical properties, they often fail to capture the context-dependent nature of molecular behavior, as a molecule’s properties can vary depending on its interactions with other molecules. To address this limitation, we employ embeddings generated by ChemBERTa, a Masked Language Model (MLM) trained exclusively on SMILES representations \cite{chithrananda2020chemberta}. These embeddings encode rich chemical information, including semantic relationships, substructure patterns, and functional group interactions, learned through extensive self-supervised pretraining on large-scale chemical language corpora.

For each SMILES input, the tokenized sequence is processed by the pretrained transformer stack and a fixed-size molecular embedding was extracted from the last hidden layer. ChemBERTa was used as a frozen feature extractor and its outputs were standardized with the same normalization operator used for descriptors.


\subsubsection{Graph Representation}

Molecules can be naturally represented as graphs, where atoms are modeled as nodes and bonds as edges. To effectively leverage the spatial and relational characteristics of molecular structures, we employed graph-based features as one of the primary inputs to our model. We utilized DGL-LifeSci toolkit \cite{dgllife} to generate these graph representations, which converts SMILES strings into molecular graphs.
The toolkit's featurizers generate a set of numerical attributes for both nodes and edges, capturing key chemical properties that can be directly used by Graph Neural Networks (GNNs).

For baseline models, we predominantly utilize the Canonical Featurizer, which produces traditional, handcrafted feature vectors that encode chemically relevant information. However, where applicable, model-specific featurizers (such as, the AttentiveFP featurizer) were employed. AttentiveFP employs its own specialized featurizer, which learns embeddings that capture complex interaction patterns not explicitly represented in the canonical descriptors. These learned features enable AttentiveFP to potentially model richer, more intricate molecular behaviors.

\subsection{Proposed Architecture}
\begin{figure}[htb]
        \centering
        \includegraphics[width=\textwidth]{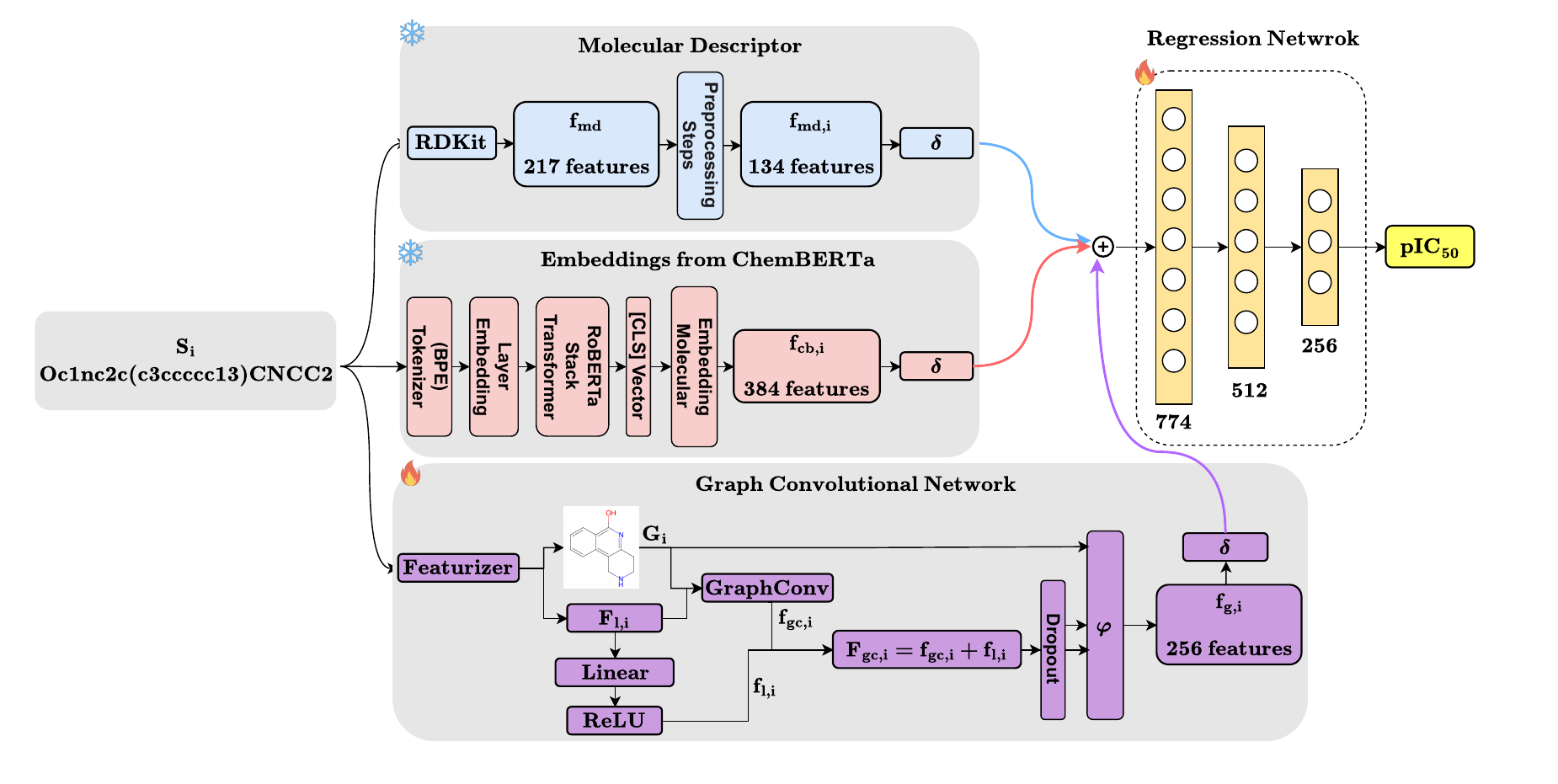}
\caption{Overview of Rep3Net architecture. Molecular descriptors, embeddings from ChemBERTa, and graph convolutional features are extracted in parallel and fused via feature concatenation. The combined representation is processed by a regression network to produce the final $pIC_{50}$ prediction.}
\label{figure: architecture}
\end{figure}

Our proposed, Rep3Net integrates the three modalities described above into a single predictive model. The architecture comprises frozen and trainable modules: RDKit descriptor computation and ChemBERTa embedding extraction are treated as frozen feature extractors, whereas the graph convolutional module and the regression head are trained on PARP1. For each compound, the three representations are computed in parallel, concatenated, and then passed to the regressor for $pIC_{50}$ prediction (\autoref{figure: architecture}).

\subsubsection{Frozen Modules}
The frozen pathways produce normalized feature vectors from RDKit and ChemBERTa, respectively, as stated in \autoref{eqn:frozen}.

\begin{equation}
\begin{aligned}
  &\hat{f}_{md,i} = \delta(\text{RDKit}(S_i)), \\
  &\hat{f}_{cb,i} = \delta(\text{ChemBERTa}(S_i))
\end{aligned}
\label{eqn:frozen}
\end{equation}

For a compound with SMILES $S_i$, the RDKit pipeline yields a descriptor vector (after feature filtering) and ChemBERTa yields a fixed-size embedding vector; both outputs are normalized with $\delta$.


\begin{equation}
\begin{aligned}
  & \delta(f) = \frac{f - \lambda}{\sigma + 1e^{-6}} \\
\end{aligned}
\label{eqn:normalization}
\end{equation}
where $\lambda$ and $\sigma$ are the mean and standard deviation of the feature vector.


\subsubsection{Graph Convolutional Network}

Graph representations $G_i$ and initial node/edge feature vectors $F_{l,i}$ are produced by the DGL-LifeSci Featurizer and processed by the GCN module of Rep3Net. The module applies graph convolution with a skip connection, dropout, and a combined pooling operator (weighted-sum and max pooling) to produce a compact graph-level feature $f_{g, i}$, which is subsequently normalized to produce $\hat{f}_{g, i}$. The computation follows \autoref{eqn:gnn}.

\begin{equation}
    \begin{aligned}
        &G_{i},F_{l,i} = \text{Featurizer}(S_{i}), \\
        &f_{gc,i} = \text{GraphConv}(G_{i},F_{l,i}), \\
        &f_{l,i} = \text{ReLU}(\text{Linear}(F_{l,i}), \\
        &F_{gc,i} = f_{gc,i} + f_{l,i}, \\
        &f_{g,i} = \varphi(\text{Dropout}(F_{gc,i}),G_{i}), \\
        &\hat{f}_{g,i} = \delta(f_{g,i})
    \end{aligned}
    \label{eqn:gnn}
\end{equation}

All graph models used in this study were initialized from weights pretrained on the BACE dataset and subsequently fine-tuned on the PARP1 data for the primary experiments and for ablation comparisons.

\subsubsection{Regression Network}
The normalized features from the three pathways, $\hat{f}_{md,i}$, $\hat{f}_{cb,i}$ and $\hat{f}_{g,i}$, are concatenated and provided to a regression head:

\begin{equation}
  \hat{Y}_{i} = \text{Regressor}(\hat{f}_{md,i} \oplus \hat{f}_{cb,i} \oplus \hat{f}_{g,i})
  \label{eqn:overall}
\end{equation} 

The regressor is a high-level multilayer perceptron comprising three fully connected layers; each layer is followed by batch normalization, a ReLU activation, and dropout to promote stable optimization and regularization (see \autoref{alg: Algo}).

\begin{algorithm}[H]{
\SetAlgoLined
\DontPrintSemicolon
\SetKwFunction{FMain}{Rep3Net}
\SetKwProg{Fn}{Function}{:}{}
\KwIn{SMILES notations of N samples, $X$.}
\KwIn{Number of epochs we will train, $EPOCHS$.}
\KwOut{Predicted $pIC_{50}$ value of N samples, $\hat{Y}$.}
\nl for $epoch$ in $EPOCHS$: \;
\nl \hspace*{1.5em} for $S_{i}$ , $Y_{i}$ in $X$: \;
\nl \hspace*{3em} $\hat{f}_{md,i}$ , $\hat{f}_{cb,i}$ , $\hat{f}_{g,i}$ $\leftarrow$ $\delta(\text{RDKit}(S_i))$, $\delta(\text{ChemBERTa}(S_i))$, $\delta(GCN(S^{i}))$ \;
\nl \hspace*{3em} $F_{c,i}$ $\leftarrow$ $(\hat{f}_{md,i} \oplus \hat{f}_{cb,i} \oplus \hat{f}_{g,i})$\;
\nl \Comment{Regression Network}\;
\nl \hspace*{3em} $x \leftarrow Dropout(ReLU(BatchNorm(FC1(F_{c,i}))))$\; 
\nl \hspace*{3em} $x \leftarrow Dropout(ReLU(BatchNorm(FC2(x))))$\;
\nl \hspace*{3em} $\hat{Y}_{i} \leftarrow FC3(x)$\;
\nl \Return $\hat{Y}$\;
}
\caption{High Level Representation of Rep3Net}
\label{alg: Algo}
\end{algorithm}




\subsection{Implementation Details}
Graph models used in Rep3Net were initialized from checkpoints pretrained on BACE and then fine-tuned on the PARP1 data. The optimization objective for all experiments is Mean Squared Error (MSE). We used Adam Optimizer with a Cosine Annealing as learning-rate scheduler. The hyperparameters used in the reported experiments were: batch size 4, learning rate $5\times10^{-5}$, weight decay $10^{-5}$, and 20 training epochs. We utilized the model that achieved the best performance over the 20 epochs.

Model selection and final evaluation were performed within each cross-validation fold using the 75:5:20 (train:val:test) partition. All reported results correspond to the test partition of each fold and are aggregated across the five folds.

For code and implementation details, see our \href{https://github.com/180041123-Atiq/Rep3Net}{project repository}\footnote{https://github.com/180041123-Atiq/Rep3Net}.

\subsection{Evaluation Metrics}
We evaluated the predictive performance of Rep3Net using six metrics that together assess error magnitude, explained variance, and rank/linear association: Mean Squared Error (MSE), Root Mean Squared Error (RMSE), Mean Absolute Error (MAE), Coefficient of Determination ($R^2$), Pearson Correlation Coefficient ($r$), and Spearman Rank Correlation Coefficient ($\rho$). All metrics were computed on the standardized $pIC_{50}$ targets.

\subsubsection{Mean Squared Error (MSE)} 
MSE quantifies the average squared difference between the predicted values $\hat{y_i}$ and the actual target values $y_i$. The values of MSE range from $0$ to $\infty$ where $0$ means the perfect prediction and higher values denote larger errors. Because the errors are squared, larger deviations are penalized more heavily.
\[
\text{MSE} = \frac{1}{n} \sum_{i=1}^{n} (y_i - \hat{y}_i)^2
\]

\subsubsection{Root Mean Squared Error (RMSE)}
RMSE is the square root of MSE, providing an interpretable measure in the same units as the target variable. Similar to MSE, RMSE ranges from $0$ to $\infty$. It remains sensitive to large prediction errors.
\[
\text{RMSE} = \sqrt{\frac{1}{n} \sum_{i=1}^{n} (y_i - \hat{y}_i)^2}
\]

\subsubsection{Mean Absolute Error (MAE)}
MAE calculates the average magnitude of prediction errors without considering direction. Unlike MSE and RMSE, it treats all deviations equally, making it robust to outliers. MAE has a range of $0$ to $\infty$, $0$ indicating a perfect prediction, and lower values mean more accurate.
\[
\text{MAE} = \frac{1}{n} \sum_{i=1}^{n} |y_i - \hat{y}_i|
\]

\subsubsection{Coefficient of Determination (\(R^{2}\))}
The \(R^{2}\) score represents the proportion of the total variance in the observed data that is explained by the model. It ranges from $-\infty$ to $1$, where $1$ indicates perfect prediction, $0$ indicates that the model has no predictive power beyond a trivial baseline and negative values indicate that the model performs worse than the baseline mean predictor.
\[
R^2 = 1 - \frac{\sum_{i=1}^{n} (y_i - \hat{y}_i)^2}{\sum_{i=1}^{n} (y_i - \bar{y})^2}
\]

where $\bar{y}$ is the mean of the actual values.

\subsubsection{Pearson Correlation Coefficient ($r$)}
Pearson Correlation evaluates the strength and direction of a linear relationship between actual and predicted values. It ranges from $-1$ to $1$, where $1$ indicates perfect positive linear correlation, $0$ indicates no linear correlation and $-1$ indicates perfect negative linear correlation.
\[
r = \frac{\sum_{i=1}^{n} (y_i - \bar{y})(\hat{y}_i - \overline{\hat{y}})}{\sqrt{\sum_{i=1}^{n} (y_i - \bar{y})^2} \sqrt{\sum_{i=1}^{n} (\hat{y}_i - \overline{\hat{y}})^2}}
\]

\subsubsection{Spearman Rank Correlation Coefficient ($\rho$)}
Spearman correlation measures the strength and direction of a monotonic (not necessarily linear) relationship. It is computed on the ranked values of the variables, making it suitable for data not following a normal distribution. The value of $\rho$ ranges from $-1$ to $1$, where $1$ indicates perfect monotonic increasing relationship, $0$ indicates no rank correlation and $-1$ indicates perfect monotonic decreasing relationship.
\[
\rho = 1 - \frac{6 \sum_{i=1}^{n} d_i^2}{n(n^2 - 1)}
\]

where $d_i$ is the difference between the ranks of $y_i$ and $\hat{y_i}$.

\section{Results and discussion}

Our objective is to predict compound $pIC_{50}$ values from structural information alone by fusing three complementary molecular representations: handcrafted molecular descriptors, learned SMILES embeddings from ChemBERTa, and graph-based representations. Experiments use fivefold cross validation with a fixed seed; reported point estimates are fold means and the accompanying ``$\pm$'' values denote the half-width of the 95\% confidence interval across folds. All metrics are computed on standardized $pIC_{50}$ targets (i.e., after \texttt{StandardScaler}). The subsections below summarize comparative performance, computational-cost trade-offs, and ablation analyses that attribute the observed gains to specific model components.

\subsection{Establishing Baseline using Classical Architectures}
Classical QSAR approaches relying exclusively on predefined molecular descriptors serve as a reference point for assessing the need for richer representations. As shown in \autoref{tbl:classical}, Random Forest, XGBoost, and KNN models perform poorly on this dataset, yielding negative or near-zero $R^2$ values. This indicates that, under the present data regime, these methods fail to capture meaningful structure–activity relationships and are outperformed by a trivial mean predictor.

\begin{table}[h]
  \caption{Performance of Standard Classical Architectures on PARP1 Inhibitors Regression Task}
  \label{tbl:classical}
  \centering
  \begin{tabular}{lllllll}
    \hline
    Model  & MSE $\downarrow$ & RMSE $\downarrow$ & MAE $\downarrow$ & $R^2$ $\uparrow$ & Pearson $\uparrow$ & Spearman $\uparrow$ \\
    \hline
    RandomForest   & $1.04\pm0.08$ & $1.02\pm0.04$ & $0.83\pm0.02$ & $-0.05\pm0.03$ & $0.02\pm0.05$ & $0.02\pm0.06$  \\
    XGBoost & $1.00\pm0.10$ & $1.00\pm0.05$ & $0.81\pm0.02$ & $-0.001\pm0.00$ & $0.01\pm0.10$ & $-0.00\pm0.12$  \\
    KNN  & $1.16\pm0.04$ & $1.08\pm0.02$ & $0.88\pm0.02$ & $-0.17\pm0.09$ & $0.02\pm0.08$ & $0.01\pm0.09$  \\
    \hline
  \end{tabular}
\end{table}

This underperformance motivated the use of learned and structured representations that can better encode subtle chemical context and nonlocal interactions. Consequently, classical models were not pursued further in this work, and the remainder of the analysis focused on neural architectures that operate directly on molecular graphs and learned embeddings.

\subsection{Comparison with GNN Baselines}
\autoref{tbl:GNN} compares the performance Rep3Net against several widely used GNN architectures. Rep3Net attains the lowest mean squared error and the highest correlation metrics among the evaluated models. Relative to individual baselines, the most salient descriptive improvements are as follows: Rep3Net reduces MSE by approximately 16.2\% relative to GAT and by approximately 20.2\% relative to Weave. For correlation metrics, Rep3Net increases $R^2$ by about 34.4\% relative to GAT and increases Pearson correlation by about 20.0\% relative to MPNN. Spearman rank correlation improves by roughly 17.5\% relative to both Weave and AttentiveFP. These effect sizes summarize where the method obtains the largest practical gains and indicate consistent, cross-fold improvements in both error and rank/linear association.

\begin{table}[h]
	\caption{Performance Comparison with GNN baselines. To ensure a fair evaluation, all baseline models were trained on the same graph representation data used by our model.}
	\label{tbl:GNN}
	\centering
	\begin{tabular}{lllllll}
		\hline
		Model  & MSE $\downarrow$ & RMSE $\downarrow$ & MAE $\downarrow$ & $R^2$ $\uparrow$ & Pearson $\uparrow$ & Spearman $\uparrow$ \\
		\hline
		GAT \cite{veličković2018graphattentionnetworks}   & $0.99\pm0.11$ & $1.00\pm0.06$ & $0.78\pm0.02$ & $0.32\pm0.06$ & $0.57\pm0.05$ & $0.59\pm0.03$  \\
		Weave \cite{kearnes2016molecular} & $1.04\pm0.14$ & $1.02\pm0.07$ & $0.79\pm0.02$ & $0.29\pm0.06$ & $0.54\pm0.05$ & $0.57\pm0.03$  \\
		MPNN \cite{gilmer2017neuralmessagepassingquantum} & $1.03\pm0.14$ & $1.02\pm0.07$ & $0.78\pm0.03$ & $0.30\pm0.05$ & $0.55\pm0.05$ & $0.55\pm0.03$  \\
		AttentiveFP \cite{xiong2019pushing} & $1.04\pm0.14$ & $1.02\pm0.07$ & $0.80\pm0.03$ & $0.29\pm0.05$ & $0.54\pm0.05$ & $0.57\pm0.04$ \\
		\textbf{Ours} & $\textbf{0.83} \pm \textbf{0.06}$ & $\textbf{0.91} \pm \textbf{0.03}$ & $\textbf{0.69} \pm \textbf{0.02}$ & $\textbf{0.43} \pm \textbf{0.01}$ & $\textbf{0.66}\pm\textbf{0.01}$ & $\textbf{0.67}\pm\textbf{0.01}$ \\
		\hline
	\end{tabular}
\end{table}

Among the baselines, GAT \cite{veličković2018graphattentionnetworks} applies attention over local neighborhoods to weight neighboring atoms during message passing. While effective in many molecular tasks, its reliance on local aggregation limits its ability to capture broader chemical context when training data are scarce. In contrast, Rep3Net achieves a substantially higher $R^2$ and correlation scores, indicating improved sensitivity to $pIC_{50}$ variation.

Weave \cite{kearnes2016molecular} employs shallow message passing focused on local chemical environments. This design restricts its ability to model long-range dependencies and global molecular properties. By integrating global descriptors, contextual ChemBERTa embeddings, and graph-based features, Rep3Net provides a more complete representation of molecular structure, enabling it to capture subtle structure-activity signals that Weave cannot.

MPNN \cite{gilmer2017neuralmessagepassingquantum} iteratively aggregates messages along graph edges, making performance strongly dependent on graph topology and predefined atom and bond features. While effective for encoding local connectivity, this approach lacks explicit mechanisms to incorporate broader chemical semantics. Rep3Net addresses this limitation by jointly leveraging graph representations and sequence-based embeddings, resulting in improved correlation with experimental $pIC_{50}$ values.

AttentiveFP \cite{xiong2019pushing} introduces hierarchical attention at both atom and molecule levels, allowing the model to emphasize salient substructures. However, this attention-heavy design is prone to overfitting and reduced generalization in low-data regimes. Rep3Net achieves a higher rank correlation, suggesting that combining complementary representations is more effective than relying solely on attention-based graph reasoning under these conditions.

In summary, Rep3Net yields substantial effect sizes relative to all GNN baselines. These improvements are consistent across folds and metrics, underscoring that the gains reflect meaningful reductions in prediction error rather than marginal metric fluctuations. These gains can be attributed to the multimodal fusion strategy. Standalone graph models rely primarily on topological message passing and node/edge features, which can be limited when the training set is small. By contrast, Rep3Net supplements graph-derived structure with global physico-chemical descriptors and contextual SMILES embeddings from a pretrained transformer, enabling the model to exploit signals that are complementary to local message passing. The GNN baselines remain valuable comparators; however, the multimodal fusion yields both lower error and stronger correlation with experimental measurements across folds.

\subsection{Computational Cost Analysis}
\begin{figure}[h]
	\centering
	\includegraphics[width=.8\textwidth]{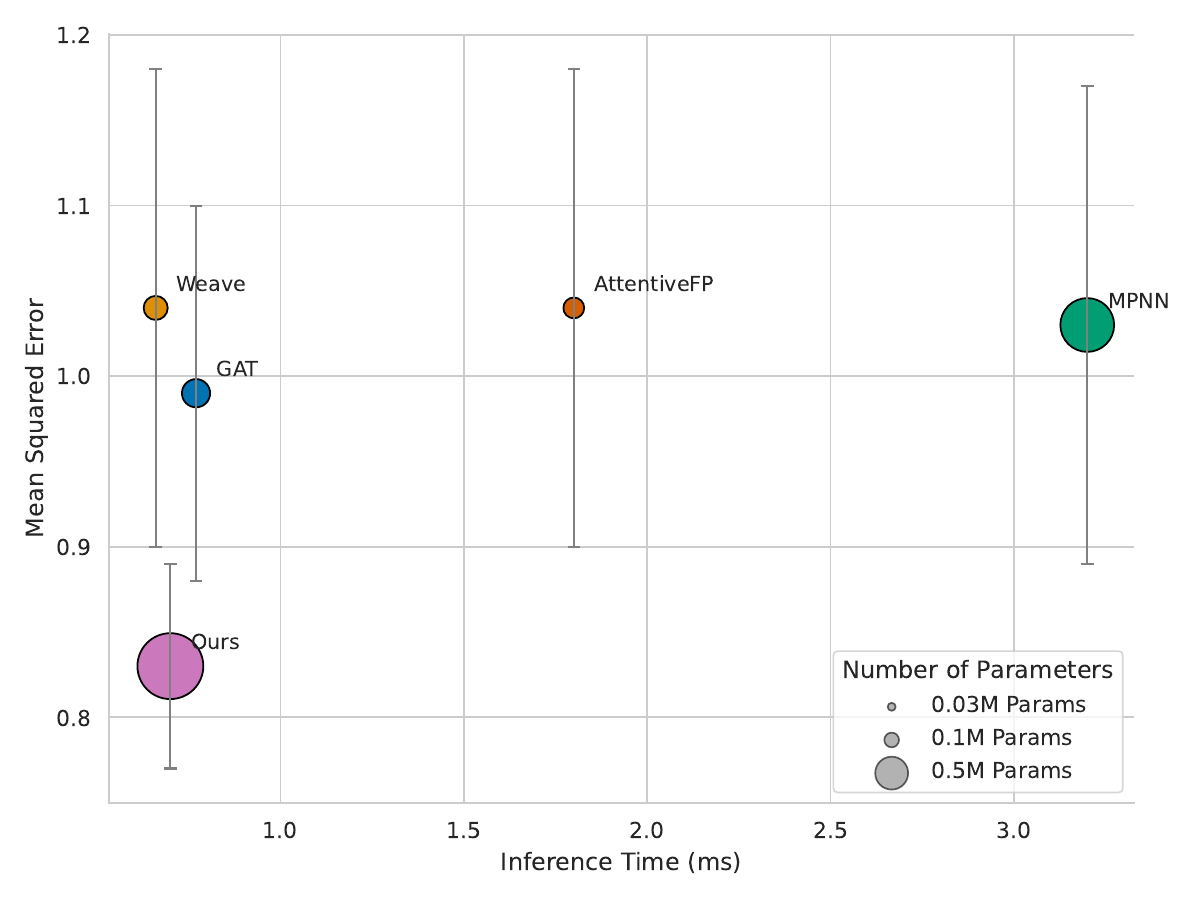}
	\caption{Trade-off between computational efficiency and predictive accuracy 
		for GNN models. The x-axis shows inference time (lower is better), and 
		the y-axis shows mean squared error (lower is better). Error bars represent 
		the half-width of the 95\% confidence interval across folds. Marker size 
		corresponds to the number of trainable parameters (in millions). Our method 
		strikes the best balance with the lowest error and competitive inference 
		efficiency.}
	\label{tbl:computation}
\end{figure}


\autoref{tbl:computation} compares the GNN models across prediction accuracy (MSE), model size, and inference speed. Among all methods, Rep3Net achieves the best predictive performance with the lowest MSE (0.83) and the tightest 95\% confidence interval ($\pm$0.06), indicating both higher accuracy and greater stability. The baseline models (GAT, Weave, MPNN, and AttentiveFP) have similar MSE values clustered around 1.0, with wider and comparable confidence intervals ($\pm$0.11-0.14). In terms of model size, Rep3Net has the largest number of parameters (0.55M), followed by MPNN (0.35M), while AttentiveFP is the smallest (0.01M). Despite its larger size, Rep3Net remains computationally efficient, with fast inference time (0.7ms), comparable to GAT (0.77ms) and Weave (0.66ms), and substantially faster than MPNN (3.2ms) and AttentiveFP (1.8ms). These results suggest that, although inference latency is partly influenced by model size, architectural design choices have a comparable impact on runtime performance.

Weave and GAT exhibit fast inference due to their relatively simple forward passes, which rely on computationally inexpensive neighborhood aggregation. AttentiveFP, despite having the fewest parameters, incurs higher latency due to stacked attention layers and sequential readout operations that introduce nontrivial overhead. MPNN shows the highest inference time, reflecting the cost of repeated message-passing steps and graph traversals.

To better contextualize efficiency, we consider inference latency normalized by model size. When measured in milliseconds per million trainable parameters, Rep3Net achieves approximately 1.27 ms per $10^6$ parameters, which is substantially lower than MPNN (9.14), GAT (12.83), Weave (22.00), and AttentiveFP (180.00). This heuristic comparison indicates that, for the reported measurements, Rep3Net delivers a favorable accuracy-efficiency trade-off relative to its parameter budget.

This efficiency arises from three architectural choices in Rep3Net: the use of a single graph convolution layer with a canonical featurizer, parallel generation of frozen ChemBERTa and descriptor-based embeddings, and the absence of deep attention stacks or iterative message passing. Together, these design elements enable scalable inference while maintaining strong predictive performance.

\subsection{Ablation Study}

\autoref{tbl:ablation} presents an ablation study that systematically isolates each input modality and evaluates their pairwise and full combinations to quantify their individual and joint contributions.

\begin{table}[h]
  \caption{Ablation study analyzing the impact of individual and combined feature representations on predictive performance.}
  \label{tbl:ablation}
  \centering
  \begin{tabular}{C{2cm} C{2cm} C{2cm}|C{1.8cm} C{2cm}}
    \hline
    Molecular Descriptor  & ChemBERTa Embeddings & Graph Convolutional Network & MSE $\downarrow$ & Spearman  $\uparrow$ \\
    \hline
    \checkmark & & & 1.06 & 0.58 \\
     & \checkmark & & 0.94 & 0.63  \\
    & & \checkmark & 0.90 & 0.66 \\
    \checkmark & \checkmark & & 0.91 & 0.65 \\
     & \checkmark & \checkmark & 0.89 & 0.66 \\
     \checkmark & & \checkmark & 0.89 & 0.67 \\
     \checkmark & \checkmark & \checkmark & \textbf{0.84} & \textbf{0.68} \\
    \hline
  \end{tabular}
\end{table}

Among individual inputs, graph-based representations achieve the strongest performance (MSE = 0.90), outperforming both ChemBERTa embeddings (MSE = 0.94) and handcrafted molecular descriptors (MSE = 1.06). This ordering suggests that explicit molecular topology provides the most informative standalone signal for structure–activity prediction, while contextual SMILES semantics capture richer cues than global descriptors alone. The comparatively weaker descriptor-only performance indicates that handcrafted global statistics do not fully encode the fine-grained structural features required for accurate activity modeling.

Combining any two modalities consistently reduces MSE and improves Spearman rank correlation relative to single-modality models. These gains demonstrate clear complementarity between representations: each modality captures partially distinct aspects of molecular structure and activity, and their integration yields more accurate and better-ranked predictions.

The model that fuses descriptors, ChemBERTa embeddings, and graph topology achieves the best overall performance (MSE = 0.84; Spearman = 0.68), confirming that all three modalities contribute additive information. Quantitatively, relative to the full model, descriptors-only exhibits a 20.8\% higher MSE, ChemBERTa-only a 10.6\% higher MSE, and graph-only a 6.7\% higher MSE. Moreover, extending the strongest two-modality configuration (MSE = 0.89) to the full fusion yields an additional $\approx$5.6\% relative reduction in MSE, indicating diminishing but consistent marginal gains as modalities are added.

These trends support a mechanistic view of modality complementarity: global descriptors encode coarse physico-chemical priors, ChemBERTa embeddings capture substructure-aware contextual semantics learned from large SMILES corpora, and graph representations encode local bonding patterns and topology that are critical for resolving activity cliffs. Their fusion produces the most accurate and stable predictions on the evaluated PARP1 subset of ChEMBL, highlighting the value of multimodal integration for molecular property prediction.

\section{Conclusion}
We present Rep3Net, a multimodal deep learning framework that integrates RDKit molecular descriptors, pretrained SMILES embeddings from ChemBERTa, and graph-based structural representations within a residual GCN–regressor architecture to predict $pIC_{50}$ for PARP1. On a curated ChEMBL PARP1 dataset, Rep3Net consistently outperforms classical descriptor-only models and several recent GNN baselines, achieving an MSE of $0.83\pm0.06$ and substantial relative reductions in prediction error ($\approx$16–20\% versus individual GNNs). The ablation study indicates that each modality supplies nonredundant information: graph topology provides the strongest single-modality rank signal, while descriptors and ChemBERTa embeddings add complementary global and contextual priors that further reduce error when fused. Architecturally, Rep3Net attains a favorable accuracy-efficiency balance by combining a streamlined graph component with parallel, frozen feature extractors, enabling practical inference throughput for virtual screening workflows. While the present evaluation focuses on PARP1 within the ChEMBL chemical space, the approach is broadly applicable: the modular design supports additional molecular modalities, larger chemical libraries, and transfer to other targets. Future work will explore wider external validation, improved interpretability for lead optimization, and cross-target transfer learning to extend Rep3Net's utility across drug-discovery campaigns.

\printbibliography




  
  

\end{document}